\documentclass{article}
\usepackage{amsmath, amssymb, amsthm}
\usepackage{mathtools}
\usepackage{bbm}
\usepackage{dsfont}

\usepackage[usenames,dvipsnames]{xcolor}

\usepackage{enumitem}

\usepackage{graphicx}
\usepackage[labelfont=bf]{caption}
\usepackage[format=hang]{subcaption}

\usepackage{booktabs, array}
\usepackage{multirow}
\usepackage{subcaption}

\newsavebox\CBox

\newcommand{\tbf}[1]{\sbox\CBox{#1}\resizebox{\wd\CBox}{\ht\CBox}{\textbf{#1}}}
\newcommand{\tpm}[1]{\small{$\pm{\, #1}$}}
\newcommand{\umet}[1]{#1 ($\uparrow$)}
\newcommand{\dmet}[1]{#1 ($\downarrow$)}

\usepackage{ifthen}
\newcommand{\tv}[3][n]{%
    \ifthenelse{\equal{#1}{n}}{%
        \ifthenelse{\equal{#3}{}}{#2}{#2 \tpm{#3}}%
    }{\ifthenelse{\equal{#1}{b}}{%
        \ifthenelse{\equal{#3}{}}{\tbf{#2}}{\tbf{#2} \tpm{#3}}%
    }{\ifthenelse{\equal{#1}{g}}{%
        \ifthenelse{\equal{#3}{}}{\color{gray}#2}{\color{gray}#2 \tpm{#3}}%
    }{%
        \textcolor{red}{ERROR}%
    }}%
}}

\usepackage{listings}
\usepackage{fancyvrb}
\fvset{fontsize=\small}

\usepackage{natbib}
\usepackage[colorlinks,linktoc=all,pagebackref=true]{hyperref}
\usepackage[hypcap=true]{caption}
\hypersetup{citecolor=NavyBlue}
\hypersetup{linkcolor=NavyBlue}
\hypersetup{urlcolor=NavyBlue}
\usepackage[nameinlink,capitalise]{cleveref}
\creflabelformat{equation}{#2\textup{#1}#3}  % <- remove parenthesis from equations

\lstdefinestyle{mystyle}{
    commentstyle=\color{OliveGreen},
    numberstyle=\tiny\color{black!60},
    stringstyle=\color{BrickRed},
    basicstyle=\ttfamily\scriptsize,
    breakatwhitespace=false,
    breaklines=true,
    captionpos=b,
    keepspaces=true,
    numbers=none,
    numbersep=5pt,
    showspaces=false,
    showstringspaces=false,
    showtabs=false,
    tabsize=2
}
\usepackage[acronym,nowarn,section,nogroupskip,nonumberlist]{glossaries}
\glsdisablehyper{}

\newacronym{mlp}{\textsc{mlp}}{Multilayer perceptron}
\newcommand{\bz}{\mathbf{z}}

\newcommand{\bsa}{\boldsymbol{a}}
\newcommand{\bsb}{\boldsymbol{b}}

\newcommand{\bsh}{\boldsymbol{h}}

\newcommand{\bsm}{\boldsymbol{m}}

\newcommand{\bsr}{\boldsymbol{r}}
\newcommand{\bss}{\boldsymbol{s}}

\newcommand{\bsx}{\boldsymbol{x}}
\newcommand{\bsy}{\boldsymbol{y}}
\newcommand{\bsz}{\boldsymbol{z}}

\newcommand{\bsK}{\boldsymbol{K}}

\newcommand{\bsU}{\boldsymbol{U}}

\newcommand{\bsW}{\boldsymbol{W}}

\newcommand{\calL}{{\mathcal{L}}}

\newcommand{\calN}{{\mathcal{N}}}

\newcommand{\bbE}{\mathbb{E}}

\newcommand{\bbR}{\mathbb{R}}

\newcommand{\bgamma}{{\boldsymbol{\gamma}}}
\newcommand{\bdelta}{{\boldsymbol{\delta}}}

\newcommand{\btheta}{{\boldsymbol{\theta}}}

\newcommand{\blambda}{{\boldsymbol{\lambda}}}
\newcommand{\bmu}{{\boldsymbol{\mu}}}

\newcommand{\bsigma}{{\boldsymbol{\sigma}}}

\newcommand{\bphi}{{\boldsymbol{\phi}}}

\newcommand{\bpsi}{{\boldsymbol{\psi}}}

\theoremstyle{plain}% default

\theoremstyle{definition}

\theoremstyle{remark}

\newcommand{\dee}{\mathrm{d}}

\newcommand{\tr}{^\top}

\newcommand{\iidsim}{\overset{\mathrm{i.i.d.}}{\sim}}

\def\[#1\]{\begin{align}#1\end{align}}

\usepackage{accents}
\newcommand{\obs}{^{\text{o}}}
\newcommand{\mis}{^{\text{m}}}

\usepackage{microtype}
\usepackage{graphicx}
\usepackage{booktabs} % for professional tables

\usepackage{hyperref}

\usepackage[accepted]{icml2023}

\usepackage{amsmath}
\usepackage{amssymb}
\usepackage{mathtools}
\usepackage{amsthm}

\theoremstyle{plain}

\theoremstyle{definition}

\theoremstyle{remark}

\usepackage[textsize=tiny]{todonotes}

\icmltitlerunning{Probabilistic Imputation for Time-series Classification with Missing Data}

\begin{document}

\twocolumn[
\icmltitle{Probabilistic Imputation for Time-series Classification with Missing Data}



\icmlsetsymbol{equal}{*}

\begin{icmlauthorlist}
\icmlauthor{SeungHyun Kim}{equal,kaist}
\icmlauthor{Hyunsu Kim}{equal,kaist}
\icmlauthor{EungGu Yun}{equal,kaist,saige}
\icmlauthor{Hwangrae Lee}{samsung}
\icmlauthor{Jaehun Lee}{samsung}
\icmlauthor{Juho Lee}{kaist,aitrics}
\end{icmlauthorlist}

\icmlaffiliation{kaist}{Kim Jaechul Graduate School of AI, Korea Advanced Institute of Science and Technology (KAIST), Daejeon, South Korea}
\icmlaffiliation{saige}{Saige Research, Seoul, South Korea}
\icmlaffiliation{samsung}{Samsung Research, Seoul, South Korea}
\icmlaffiliation{aitrics}{AITRICS, Seoul, South Korea}

\icmlcorrespondingauthor{SeungHyun Kim}{sh0776879@gmail.com}
\icmlcorrespondingauthor{Juho Lee}{juholee@kaist.ac.kr}

\icmlkeywords{Machine Learning, ICML}

\vskip 0.3in
]



\printAffiliationsAndNotice{\icmlEqualContribution} % otherwise use the standard text.

\begin{abstract}

Multivariate time series data for real-world applications typically contain a significant amount of missing values. The dominant approach for classification with such missing values is to impute them heuristically with specific values (zero, mean, values of adjacent time-steps) or learnable parameters. However, these simple strategies do not take the data generative process into account, and more importantly, do not effectively capture the uncertainty in prediction due to the multiple possibilities for the missing values. In this paper, we propose a novel probabilistic framework for classification with multivariate time series data with missing values. Our model consists of two parts; a deep generative model for missing value imputation and a classifier. Extending the existing deep generative models to better capture structures of time-series data, our deep generative model part is trained to impute the missing values in multiple plausible ways, effectively modeling the uncertainty of the imputation. The classifier part takes the time series data along with the imputed missing values and classifies signals, and is trained to capture the predictive uncertainty due to the multiple possibilities of imputations. Importantly, we show that na\"ively combining the generative model and the classifier could result in trivial solutions where the generative model does not produce meaningful imputations. To resolve this, we present a novel regularization technique that can promote the model to produce useful imputation values that help classification. Through extensive experiments on real-world time series data with missing values, we demonstrate the effectiveness of our method.

\end{abstract}

\section{Introduction}
\label{main:sec:introduction}
Multivariate time-series data are universal; many real-world applications ranging from healthcare, stock markets, and weather forecasting take multivariate time-series data as inputs. Arguably the biggest challenge in dealing with such data is the presence of missing values, due to the fundamental difficulty of faithfully measuring data for all time steps. The degree of missing is often severe, so in some applications, more than 90\% of data are missing for some features. Therefore, developing an algorithm that can accurately and robustly perform predictions with missing data is considered an important problem to be tackled.

In this paper, we focus on the task of classification, where the primary goal is to classify given multivariate time-series data with missing values, simply imputing the missing values with heuristically chosen values considered to be strong baselines that are often competitive or even better than more sophisticated methods. For instance, one can fill all the missing values with zero, the mean of the data, or values from the previous time steps. GRU-D~\citep{che2018recurrent} proposes a more elaborated imputation algorithm where the missing values are filled with a mixture between the data means and values from the previous time steps with the mixing coefficients learned from the data. While these simple imputation-based methods work surprisingly well~\citep{che2018recurrent,du2022saits}, they lack a fundamental mechanism to recover the missing values, especially the underlying generative process of the given time series data. 

Dealing with missing data is deeply connected to handling uncertainties originating from the fact that there may be multiple plausible options for filling in the missing values, so it is natural to analyze them with the probabilistic framework. There have been rich literature on statistical analysis for missing data, where the primary goal is to understand how the observed and missing data are generated. In the seminal work of \citet{little2002statistical}, three assumptions for the missing data generative process were introduced, including Missing Completely At Random (MCAR), Missing At Random (MAR), and Missing Not At Random (MNAR). While MCAR or MAR simplifies the modeling and thus makes the inference easier, they may be unrealistic for real-world applications, because they assume that the missing mechanism is independent of the missing values (MAR) or both missing and observed values (MCAR). MNAR, the most generic assumption, assumes that the missing mechanism depends on both missing and observed values, so the generative model based on the MNAR assumption should explicitly take the missing mechanism into account. Based on this framework, \citet{mattei2019miwae} presented deep generative models for missing data under MAR assumption, and this was later extended to MNAR in \citet{ipsen2021notmiwae}. Combining the deep generative model and classifier, \citet{ipsen2022how} proposed a hybrid model that can classify missing data with problematically imputed values generated under MAR assumption.

Still, in our opinion, there is no satisfactory work combining probabilistic generative models for multivariate time-series data with missing values and classification models, so that the classifier could consider the uncertainty in filling in the missing values when making predictions. The aforementioned probabilistic frameworks are not designed for classification~\citep{mattei2019miwae,ipsen2021notmiwae}, and more importantly, not tailored for time series data~\citep{ipsen2022how}. A na\"ive extension of \citet{ipsen2022how} for time series is likely to fail; putting the obvious difference between the static and time series data aside, the fundamental difficulty of learning the generative models for missing is that there are no explicit learning signals that could promote the model to generate ``meaningful'' missing values. Since we don't have ground truth for the missing values, in principle, the generative model can generate arbitrary values (e.g., zeros), and the combined classifier can still successfully classify time series data, which is a critical problem that is overlooked in the existing works.

To this end, we propose a hybrid model combining the deep generative models for multivariate time series data and the classification models for them. The generative part is built under the MNAR assumption and is designed to naturally encode the continuity of the multivariate time series data. The classifier then takes the missing values generated from the generative model to classify time-series, and unlike the algorithms based on heuristic imputations, it takes multiple feasible options for the missing values and computes predictions based on them. To tackle the difficulty in guiding the generative model to generate ``meaningful" missing values, we introduce a novel regularization technique that deliberately erases \emph{observed values} during training. As a consequence, the classifier is forced to do classification based more on the generated missing values,  so the generative model is encouraged to produce missing values that are more advantageous for the classification. Using the various real-world multivariate time series benchmarks with missing values, we demonstrate that our approach outperforms baselines both in terms of classification accuracy and uncertainty estimates.

\section{Background}
\label{main:sec:background}
\subsection{Settings and Notations}
Let $\bsx = [x_1,\dots, x_d]\tr \in \bbR^d$ be a $d$-dimensional vector, along with the mask vector $\bss = [s_1,\dots, s_d]\tr \in \{0,1\}^d$, where $s_j=1$ if $x_j$ is observed and $s_j=0$ otherwise.  Given a mask $\bss$, we can split $\bsx$ into the observed part $\bsx\obs := \{ x_j\,|\, s_j=1\}$ and the missing part $\bsx\mis := \{x_j\,|\, s_j=0\}$. For a collection of data, the $i^\text{th}$ instance is denoted as $\bsx_i = [x_{i,1},\dots, x_{i,d}]$, and $\bss_i$, $\bsx\obs_i$, and $\bsx\mis_i$ are defined similarly. For a multivariate time-series data, we denote the vector of $t^\text{th}$ time step as $\bsx_t = [x_{t,1},\dots, x_{t,d}] \in \bbR^d$, and the corresponding mask as $\bss_t = [s_{t,1},\dots, s_{t,d}]$. The $t^\text{th}$ time step of $i^\text{th}$ instance of a collection is denoted as $\bsx_{t,i}$, which is split into $\bsx_{t,i}\obs$ and $\bsx_{t,i}\mis$ according to $\bss_{t,i}$. 

Following \citet{mattei2019miwae,ipsen2021notmiwae}, we assume that the joint distribution of an input $\bsx$ and a mask $\bss$ is factorized as $p_{\btheta, \bpsi}(\bsx, \bss) = p_\btheta(\bsx) p_\bpsi(\bss|\bsx)$. The conditional distribution $p_\bpsi(\bss|\bsx)$ plays an important role for describing missing mechanism. Under MCAR assumption, we have $p(\bss|\bsx) = p(\bss)$, under MAR we have $p_\bpsi(\bss|\bsx)=p_\bpsi(\bss|\bsx\obs)$, and under MNAR we have $p_\bpsi(\bss|\bsx) = p_\bpsi(\bss|\bsx\obs,\bsx\mis)$. The likelihood for the observed data $\bsx\obs$ is thus computed as $p_{\btheta,\bpsi}(\bsx\obs, \bss) = \int p_{\btheta,\bpsi}(\bsx, \bss) \dee \bsx\mis$.

\subsection{Missing Data Importance-Weighted Autoencoder and its extensions}
In this section, we briefly review the Missing data Importance-Weighted AutoEncoder (MIWAE)~\citep{mattei2019miwae}, a deep generative model for missing data, and its extensions to MNAR and supervised settings. Similar to variational autoencoder (VAE)~\citep{kingma2014auto}, MIWAE assumes that a data $\bsx$ is genearted from a latent representation $\bsz$, but we only observe $\bsx\obs$ with $\bss$ generated from the missing model $p_{\bpsi}(\bss|\bsx)$. MIWAE assumes MAR, so we have $p_{\bpsi}(\bss|\bsx) = p_{\bpsi}(\bss|\bsx\obs)$. The log-likelihood for $(\bsx\obs,\bss)$ is then computed as
\[
\lefteqn{\log p_{\btheta,\bpsi}(\bsx\obs,\bss)}\nonumber\\
% &= \log\int p_{\bpsi}(\bss|\bsx\obs) p_\btheta(\bsx\obs, \bsx\mis|\bsz) p_\btheta(\bsz) \dee \bsz \dee \bsx\mis \nonumber\\
&= \log p_\bpsi(\bss|\bsx\obs) + \underbrace{\log \int p_\btheta(\bsx\obs|\bsz) p_\btheta(\bsz)\dee \bsz}_{=\log p_\btheta(\bsx\obs)}. 
\]
For the missing data imputation, $p_\bpsi(\bss|\bsx\obs)$ is not necessary, so we choose to maximize only the $\log p_\btheta(\bsx\obs)$. The integral is intractable, so we consider the Importance Weighted AutoEncoder (IWAE) lower bound~\citep{burda2015iwae} as a proxy loss,
\[
\calL^{(K)}_{\text{MIWAE}}(\btheta,\bphi) :=
\bbE\bigg[\log \frac{1}{K}\sum_{k=1}^K \frac{p_\btheta(\bsx\obs|\bsz_k)p_\btheta(\bsz_k)}{q_\bphi(\bsz_k|\bsx\obs)}\bigg].
\]
Here, $q_\bphi(\bsz_k|\bsx\obs)$ for $k=1,\dots, K$ are i.i.d. copies of the variational distribution (encoder) $q_\bphi(\bsz|\bsx\obs)$ approximating the true posterior $p_\bphi(\bsz|\bsx\obs)$, and the expectation is w.r.t. $\prod_{k=1}^K q(\bsz_k|\bsx\obs)$. $\bbE_{\bsz_{1:K}}$ denotes the expectation w.r.t. $\prod_{k=1}^K q_\bphi(\bsz_k|\bsx\obs)$. $K$ is the number of particles, and the bound converges to the log-likelihood as $K\to\infty$, that is, $\calL^{(1)}_\text{MIWAE}(\btheta,\bphi) \leq \calL^{(2)}_\text{MIWAE}(\btheta,\bphi) \leq \dots = \log p_\btheta(\bsx\obs)$.

\citet{ipsen2021notmiwae} presented not-MIWAE, an extension of MIWAE with MNAR assumption. The log-likelihood for $(\bsx\obs,\bss)$ under the MNAR assumption is,
\[
\log p_{\btheta,\bpsi}(\bsx\obs,\bss) &= \log \int p_\bpsi(\bss|\bsx\obs,\bsx\mis) p_\btheta(\bsx\obs|\bsz)\nonumber\\
& \times p_\btheta(\bsx\mis|\bsz) p_\btheta(\bsz)\dee\bsz\dee\bsx\mis,
\]
where we assume that $(\bsx\obs,\bsx\mis)$ are independent given $\bsz$. The corresponding IWAE lower-bound with the  variational distribution $q_\bphi(\bsx\mis, \bsz|\bsx\obs) = p_\btheta(\bsx\mis|\bsz) q_\bphi(\bsz|\bsx\obs)$ is,
\[
\lefteqn{\calL_\text{notMIWAE}^{(K)}(\btheta,\bpsi,\bphi)}\nonumber\\
&:= \bbE
\bigg[ \log \frac{1}{K}\sum_{k=1}^K \frac{p_\btheta(\bss|\bsx\obs,\bsx\mis_k)p_\btheta(\bsx\obs|\bsz_k)p_\btheta(\bsz_k)}{q_\bphi(\bsz_k|\bsx\obs)}\bigg], 
\]
where the expectation is w.r.t. $\prod_{k=1}^K p_\btheta(\bsx\mis_k|\bsz_k)q_\bphi(\bsz_k|\bsx\obs)$.

On the other hand, \citet{ipsen2022how} extended MIWAE to a supervised learning setting, where the goal is to learn the joint distribution
of an observed input $\bsx\obs$, a mask $\bss$, and corresponding label $\bsy$,
\[
\lefteqn{\log p_{\btheta,\bpsi,\blambda}(\bsy,\bsx\obs, \bss)= \log p_\bpsi(\bss|\bsx\obs) }\nonumber\\
% &= \log \int p_\blambda(\bsy|\bsx\obs, \bsx\mis)  p_\bpsi(\bss|\bsx\obs,\bsx\mis) p_\btheta(\bsx\obs, \bsx\mis|\bsz) p_\btheta(\bsz)\dee\bsz \dee\bsx\mis\nonumber\\
& + \underbrace{\log \int p_\blambda(\bsy|\bsx\obs, \bsx\mis) p_\btheta(\bsx\obs|\bsz) p_\btheta(\bsx\mis|\bsz) p_\btheta(\bsz)\dee\bsz}_{=\log p_{\btheta,\blambda}(\bsy, \bsx\obs)},
\]
The term $p_\bpsi(\bss|\bsx\obs)$ is irrelevant to the prediction for $\bsy$, so we choose to maximize $\log p_{\btheta,\blambda}(\bsy,\bsx\obs)$, which again can be lower-bounded by IWAE bound with the variational distribution $q_\bphi(\bsz,\bsx\mis|\bsx\obs) = p_\btheta(\bsx\mis|\bsz) q_\bphi(\bsz|\bsx\obs)$:
\[
\lefteqn{\calL^{(K)}_\text{supMIWAE}(\btheta,\blambda,\bphi)}\nonumber\\
&:= {\bbE}\bigg[\log \frac{1}{K}\sum_{k=1}^K \frac{p_\blambda(\bsy|\bsx\obs, \bsx\mis_k)p_\btheta(\bsx\obs|\bsz_k) p(\bsz_k)}{q_\bphi(\bsz_k|\bsx\obs)}\bigg],
\]
where the expectation is w.r.t. $\prod_{k=1}^K p_\btheta(\bsx\mis_k|\bsz_k)q_\bphi(\bsz_k|\bsx\obs)$.

% \[
% \calL_\mathrm{MIWAE} = \bbE_{\prod_{k=1}^K q(z_k|x^o)}\Bigg[\log\frac{1}{K}\sum_{k=1}^K\frac{p(x^o|z_k)p(z_k)}{q(z_k|x^o)}\Bigg]
% \]

% \[
% \calL_\mathrm{NotMIWAE} = \bbE_{\prod_{k=1}^Kp(x^m|z_k) q(z_k|x^o)}\Bigg[\log\frac{1}{K}\sum_{k=1}^K\frac{p(s|x^o,x^m)p(x^o|z_k)p(z_k)}{q(z_k|x^o)}\Bigg]
% \]

% \[
% \calL_\mathrm{SupMIWAE} = \bbE_{\prod_{k=1}^Kp(x^m|z_k) q(z_k|x^o)}\Bigg[\log\frac{1}{K}\sum_{k=1}^K\frac{p(y|x^o,x^m)p(x^o|z_k)p(z_k)}{q(z_k|x^o)}\Bigg]
% \]

\subsection{GRU for multivariate time series data and imputation methods}
We briefly review GRU~\citep{cho2014learning} and its variant for time series classification with missing data since they are common baselines. Given a multivariate time series $(\bsx_t)_{t=1}^T$, GRU takes a vector of one time step at a time and accumulates the information into a hidden state $\bsh_t$. Specifically, the forward pass at  $t^\text{th}$ time step takes $\bsx_t$ and updates the hidden state $\bsh_t$ as follows:
\[
\bsa_t &= \sigma(\bsW_{\bsa}\bsx_t + \bsU_{\bsa}\bsh_{t-1} + \bsb_{\bsa}),\nonumber\\
\bsr_t &= \sigma(\bsW_{\bsr}\bsx_t + \bsU_{\bsr}\bsh_{t-1} + \bsb_{\bsr}) \nonumber\\
\tilde{\bsh}_t &= \mathrm{tanh}(\bsW\bsx_t + \bsU(\bsr_t \odot \bsh_{t-1}) + \bsb), \nonumber\\
\bsh_t &= (1-\bsa_t)\odot \bsh_{t-1} + \bsa_t\odot \tilde{\bsh}_t, \nonumber
\]
where $\odot$ denotes the element-wise multiplication. We also review the heuristical imputation methods described in \citet{che2018recurrent}, which are common baselines for the related methods.
Let $\hat{x}_{t,j}$ denote the imputed value for $x_{t,j}$.
\begin{itemize}[leftmargin=7mm]
    \item \textbf{GRU-zero}: a zero padding setting $\hat{x}_{t,j}=s_{t,j} x_{t,j}$.
    \item \textbf{GRU-mean}: imputes the missing values as $\hat{x}_{t,j} = s_{t,j} x_{t,j} + (1-s_{t,j}) \bar{x}_j$, where $\bar{x}_j = \sum_{i=1}^n \sum_{t=1}^T s_{t,i,j}x_{t,i,j}/\sum_{i=1}^n \sum_{t=1}^T s_{t,i,j}$ is the empirical mean of observed values for $j^\text{th}$ feature of a given collection of time series data $( (\bsx_{t,i})_{t=1}^T )_{i=1}^n$. 
    \item \textbf{GRU-forward}: set $\hat{x}_{t,j} = s_{t,j} x_{t,j} + (1-s_{t,j}) x_{t',j}$, where $t'$ is the last time when $j^\text{th}$ feature was observed before $t$.
    \item \textbf{GRU-simple}: along with the imputed vector $\hat{\bsx}_t$ (either by GRU-mean or GRU-forward), concatenate additional information. \citet{che2018recurrent} proposed to concatenate 1) the mask $\bss_t$, and the \emph{time-interval} $\bdelta_t$ saving the length of the intervals between observed values (see \citet{che2018recurrent} for precise definition).  The concatenated vector $[\hat{\bsx}_t, \bss_t, \bdelta_t]$ is then fed into GRU.
    \item \textbf{GRU-D}: introduces \emph{learnable decay} values for the input $\bsx_t$ and hidden state $\bsh_t$ as follows:
    \[
    \bgamma_{\bsx_t} &= \exp(-\max(\bsW_{\bgamma_{\bsx}}\bdelta_t + \bsb_{\bgamma_{\bsx}}, \mathbf{0})), \nonumber\\
    \bgamma_{\bsh_t} &= \exp(-\max(\bsW_{\bgamma_{\bsh}}\bdelta_t + \bsb_{\bgamma_{\bsh}}, \mathbf{0})).\nonumber
    \]
    Given a vector $\bsx_t$ with mask $\bss_t$, GRU-D imputes the missing values as
    \[\label{eq:grud_impute}
    \hat{x}_{t,j} &= \mathrm{Decay}(s_{t,j},x_{t,j},\gamma_{\bsx_t,j}, x_{t',j},\bar{x}_j)\nonumber\\
    &:= s_{t,j} x_{t,j} + (1-s_{t,j})(\gamma_{\bsx_t,j}x_{t',j}\nonumber\\
    & + (1-\gamma_{\bsx_t,j})\bar{x}_j).
    \]
    That is, the missing is imputed as a mixture of the last observed $x_{t',j}$ and the empirical mean $\bar{x}_j$ with the mixing coefficient set as the learned decay. The hidden state from the previous time step $\bsh_{t-1}$ is decayed as $\bgamma_{\bsh_t} \odot \bsh_{t-1}$ and passed through GRU along with the imputed $\hat{\bsx}_t$.
\end{itemize}
\section{Methods}
\label{main:sec:methods}
In this section, we describe our method, a probabilistic framework for multivariate time series data with missing values. Our method is an extension of supMIWAE to time series data under MNAR assumption, but the actual implementation is not merely a na\"ive composition of the existing models. In \cref{main:subsec:supnotmiwae}, we first present supnotMIWAE, an MNAR version of supMIWAE, with the encoder and decoder architectures designed for time series data with missing values. In \cref{main:subsec:obsdrop}, we show why the sup(not)MIWAE for data with missings may fail, and propose a novel regularization technique to prevent that.

\subsection{supnotMIWAE for multivariate time series data}
\label{main:subsec:supnotmiwae}
Given a multivariate time series data $\bsx_{1:T} := (\bsx_t)_{t=1}^T$ with observed $\bsx_{1:T}\obs$, missing $\bsx_{1:T}\mis$, a missing mask $\bss_{1:T}:=(\bss_t)_{t=1}^T$, and a label $\bsy$, we assume the following state-space model with latent vectors $\bsz_{1:T} := (\bsz_t)_{t=1}^T$.
\[
\lefteqn{p_{\btheta,\bpsi, \blambda}(\bsy, \bsx_{1:T}\obs, \bss_{1:T})}\nonumber\\
&=\int p_\blambda(\bsy|\bsx\obs_{1:T}, \bsx\mis_{1:T}) 
p_\btheta(\bsx\obs_{1:T}|\bsz_{1:T})p_\btheta(\bsx\mis_{1:T}|\bsz_{1:T})\nonumber\\
& \quad\times p_\btheta(\bz_{1:T})
p_\bpsi(\bss_{1:T}|\bsx_{1:T}) \dee \bsx\mis_{1:T} \dee \bsz_{1:T}.
\]
Below we describe each component more in detail.

% \paragraph{Prior $p_\btheta(\bsz_{1:T})$} 
% we assume an autoregressive prior for $\bsz_{1:T}$, 
% \[
% p_\btheta(\bsz_{1:T}) = \calN(\bsz_1|\boldsymbol{0}, \bsI)\prod_{t=2}^T \calN(\bsz_t | \bmu_{\text{pr}}(\bsz_{1:t-1}), \mathrm{diag}(\bsigma_{\text{pr}}^2(\bsz_{1:t-1})),
% \]
% where $(\bmu_{\text{pr}}(\bsz_{1:t}), \sigma_{\text{pr}}(\bsz_{1:t}))_{t=1}^{T-1}$ are computed as
% \[\label{eq:prior_gru}
% \bsh_t = \mathrm{GRU}_\text{pr}(\bsz_t, \bsh_{t-1}), \quad \bmu_{\text{pr}}(\bsz_{1:t}), \bsigma_{\text{pr}}(\bsz_{1:t}) = \mathrm{MLP}_\text{pr}(\bsh_{t-1}).
% \]
% Here, $\mathrm{GRU}_\text{pr}(\bsz_t, \bsh_{t-1})$ is a GRU cell that takes $\bsz_t$ and the hidden state $\bsh_{t-1}$ and update it to $\bsh_t$.
% \revise{
\paragraph{Prior $p_\btheta(\bsz_{1:T})$.}
 We assume Gaussian process prior as in \citet{forutin2020gpvae} for $\bsz_{1:T}$ to encode temporal correlation in the latent space. Let
 $\bsz_{1:T,j} = [z_{1,j}, \dots, z_{T,j}]^\top$ be the vector collecting $j^\text{th}$ dimension of the series $\bsz_{1:T}$.
 \[
p_\btheta(\bsz_{1:T}) = \prod_{j=1}^d \calN( \bsz_{1:T,j} | \mathbf{0}, \bsK),
\]
where $\bsK_{ij} =  k(t_{i},t_{j})$  $i,j \in \{1 \dots T\}$ and $k$ is kernel function. We use a Cauchy kernel for all experiments in this paper.
% \[
% p_\btheta(\bsz_{1:T}) = \prod_{t=1}^T\calN(\bsz_t|\boldsymbol{0}, \bsI).
% \]

\paragraph{Decoders $p_\btheta(\bsx\obs_{1:T}|\bsz_{1:T})$ and $p_\btheta(\bsx\mis_{1:T}|\bsz_{1:T})$.}
The decoder for the observed $p_\btheta(\bsx\obs_{1:T}|\bsz_{1:T})$ is defined in an autoregressive fashion,
\[
\lefteqn{p_\btheta(\bsx\obs_{1:T}|\bsz_{1:T})}\nonumber\\
&= \prod_{t=1}^T \calN( \bsx\obs_t | \bmu_{\text{dec}}(\bsz_{1:t}), \mathrm{diag}(\bsigma^2_{\text{dec}}(\bsz_{1:t}))),
\]
where $( \bmu_{\text{dec}}(\bsz_{1:t}), \bsigma_{\text{dec}}(\bsz_{1:t}))_{t=1}^T$ are defined with a transformer~\citep{vaswani2017attention} with causal maskings.
\[
&\bsh_t = \text{Transformer}_\text{dec}(\bsz_{1:t}), \nonumber\\
&(\bmu_{\text{dec}}(\bsz_{1:t}),\bsigma_{\text{dec}}(\bsz_{1:t})) = \mathrm{MLP}_\text{dec}(\bsh_t).
\]
In practice, this casual transformer layer is applied at times. The decoder for the missing $p_\btheta(\bsx_{1:T}\mis|\bsz_{1:T})$ defined similarly. In our implementation, we actually let them share the same model generating both $\bsx\obs_t$ and $\bsx\mis_t$.

\paragraph{Missing model $p_\bpsi(\bss_{1:T}|\bsx_{1:T})$.} The missing model is simply assumed to be independent Bernoulli distributions over the time steps and features.
\[
p_\bpsi(\bss_{1:T}|\bsx_{1:T}) = \prod_{t=1}^T \prod_{j=1}^d \mathrm{Bern}(s_{t,j}| \sigma_{\text{mis},t,j}(\bsx_{1:T})),
\] where $\sigma_{\text{mis}}(\bsx_{1:T})$ is computed as
\[
\sigma_{\text{mis}}(\bsx_{1:T}) = \mathrm{MLP}_\text{mis}(\bsx_{1:T}).
\]

\paragraph{Classifier $p_\blambda(\bsy|\bsx_{1:T}\obs, \bsx_{1:T}\mis)$} 
We use a transformer based model for the classifier. Given a time-series data $\bsx_{1:T}$ packing the observed values $\bsx_{1:T}\obs$ and the imputed missing values generated from the decoders, we first process the data with 1D CNN applied along the time axis to compute $\bsr_{1:T} := \mathrm{CNN}(\bsx_{1:T})$. Then we process $\bsr_{1:T}$
with a Transformer block to compute an output $\bsh_T$. The conditional distribution $p_\blambda(\bsy|\bsx_{1:T}\obs, \bsx_{1:T}\mis)$ is defined as
\[
\mathrm{Categorical}(\bsy \,|\, \mathrm{Softmax}(\mathrm{Linear}_\text{cls}(\bsh_T)).
\]

% \[
% &\bsh_t = \text{Transformer}_\text{cls}(\bsy|\bsx_{1:T}\obs, \bsx_{1:T}\mis), \nonumber\\
% &\mathrm{Categorical}(\bsy \,|\, \mathrm{Softmax}(\mathrm{Linear}_\text{cls}(\bsh_T)).
% \]

% We simply use a common GRU-based time series classifier for this. Let $\bsh_T$ be the hidden state from a GRU after consuming $(\bsx_{1:T}\obs, \bsx_{1:T}\mis)$. Then the conditional distribution $p_\blambda(\bsy|\bsx_{1:T}\obs, \bsx_{1:T}\mis)$ is defined as
% \[
% \mathrm{Categorical}(\bsy \,|\, \mathrm{Softmax}(\mathrm{Linear}_\text{cls}(\bsh_T)).
% \]
During the forward pass, the classifier takes the observed input $\bsx_{1:T}\obs$ and the missing values \emph{generated} from the decoder $p_{\btheta}(\bsx_{1:T}\mis|\bsz_{1:T})$. We find it beneficial to adopt the idea of GRU-D, where instead of directly putting the generated missing values $\bsx_{1:T}\mis$, putting the \emph{decayed} missing values as follows:
\[\label{eq:decayed_impute}
\tilde{\bsx}_{1:T} &:= (\bsx_{1:T}\obs, \bsx_{1:T}\mis) \text{ where } \bsx_{1:T}\mis \sim p_\btheta(\bsx_{1:T}\mis|\bsz_{1:T}),\nonumber\\
\hat{x}_{t,j} &= \mathrm{Decay}(s_{t,j},  x_{t,j}, \gamma_{\hat{\bsx}_t,j},x_{t',j}, \tilde{x}_{t,j}).
\]
where $\bgamma_{{\hat{\bsx}}_t} = \exp(-\max(\mathbf{0}, \bsW_{\hat{\bsx}}\bdelta_t + \bsb_{\hat{\bsx}})$ is a learnable decay. We find this stabilizes the learning when the generated missing values $\bsx_{1:T}\mis$ are inaccurate, for instance, in the early stage of learning. Note also the difference between \eqref{eq:decayed_impute} and the original GRU-D imputation \eqref{eq:grud_impute}. In GRU-D, the last observed values are mixed with the mean feature, while ours mix them with the generated values.

\paragraph{Encoder $q_\bphi(\bsz_{1:T}|\bsx\obs_{1:T})$.}
Given the generative model defined above, we introduce the variational distribution 
for $(\bsx\mis_{1:T}, \bsz_{1:T})$ that factorizes as,
\[
p_\btheta(\bsx\mis_{1:T}|\bsz_{1:T}) q_\bphi(\bsz_{1:T}|\bsx_{1:T}\obs).
\]
Here, the encoder $q_\bphi(\bsz_{1:T}|\bsx\obs_{1:T})$ is defined as an autoregressive model as before,
\[
\lefteqn{q_\bphi(\bsz_{1:T}|\bsx\obs_{1:T})}\nonumber\\
&= \prod_{t=1}^T \calN(\bsz_t | \bmu_\text{enc}(\bsx_{1:t}\obs), 
\mathrm{diag}(\bsigma^2_{\text{enc}}(\bsx_{1:t}\obs))).
\]
Given a series of observed values $\bsx\obs_{1:T}$, we first apply zero imputation for the missing values, that is, set $x'_{t,j} = x_{t,j}\obs$ if $s_{t,j}=1$ and $x'_{t,j}=0$ otherwise. Then we concatenate the missing indicators to $\bsx'_{1:T}$ and apply the 1D CNN to the time-axis as $\bsr_{1:T} = \mathrm{CNN}(\bsx'_{1:T})$.
Having computed $\bsr_{1:T}$, similar to the decoder, we use a transformer with causal masking to compute
\[
&\bsh_t = \text{Transformer}_\text{enc}(\bsr_{1:t}), \nonumber\\
&(\bmu_{\text{enc}}(\bsx\obs_{1:t}),\bsigma_{\text{enc}}(\bsx\obs_{1:t})) = \mathrm{MLP}_\text{enc}(\bsh_t).
\]

\paragraph{Objective.} Having all the ingredients defined, the IWAE bound for supnotMIWAE is computed as follows:
\[\label{eq:supnotmiwae_bound}
 \calL^{(K)}(\blambda,\btheta,\bpsi,\bphi)
:= \bbE\bigg[ \log \frac{1}{K} \sum_{k=1}^K \omega_k
\bigg],
\]
where the expectation is w.r.t. $K$ copies of a variational distribution,
$
\prod_{k=1}^K p_\btheta(\bsx_{k,1:T}\mis|\bsz_{k,1:T})q_\bphi(\bsz_{k,1:T}|\bsx_{1:T}\obs)
$, and $\omega_k$ is the importance weight term defined as
\[
\omega_k &:= p_\blambda(\bsy|\bsx_{1:T}\obs, \bsx\mis_{k,1:T}) p_\bpsi(\bss_{1:T}|\bsx_{1:T}\obs, \bsx\mis_{k,1:T}) \nonumber\\
& \times p_\btheta(\bsx_{1:T}\obs|\bsz_{k,1:T}) p_\btheta(\bsz_{k,1:T}) / q_\bphi(\bsz_{k,1:T}|\bsx_{1:T}\obs).
\]

\begin{figure*}
    \vspace{2mm}
    \centering
    \includegraphics[width=0.97\linewidth]{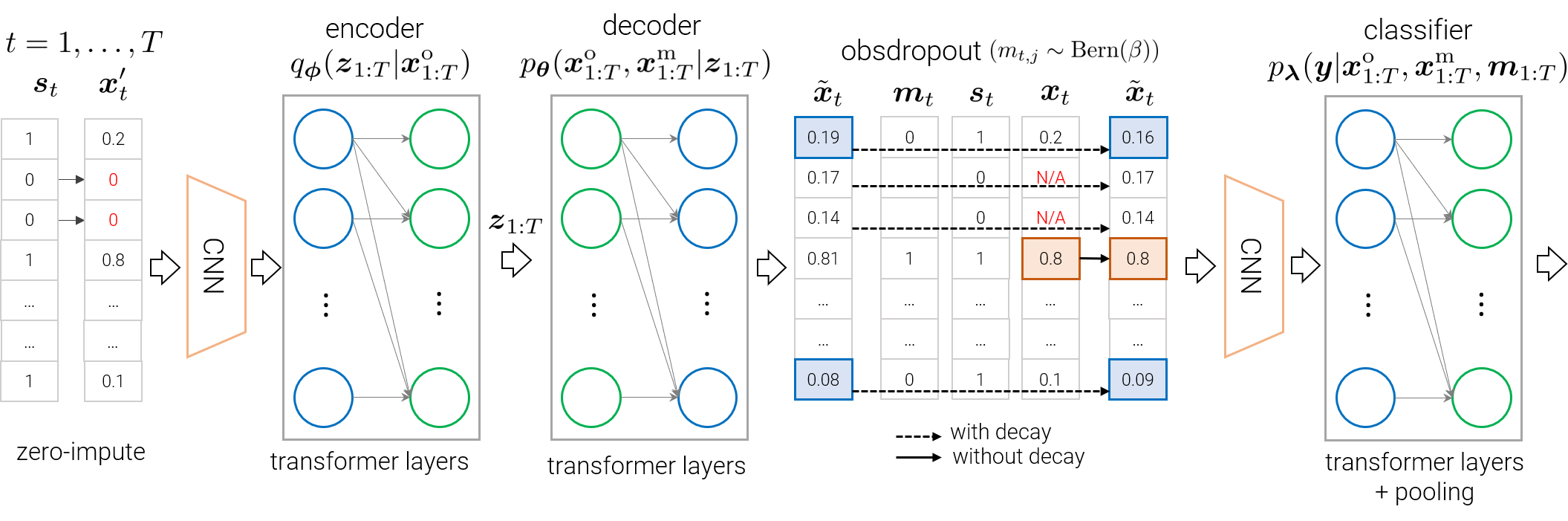}
    \caption{An overview of our model with obsdropout.}
    \label{fig:overview}
\end{figure*}

\subsection{ObsDropout: regularizing supnotMIWAE for better imputation}
\label{main:subsec:obsdrop}
The problem with \eqref{eq:supnotmiwae_bound} is that there is no clear supervision for the missing values $\bsx\mis_{1:T}$.
Obviously, if we had an access to the missing values, the conditional probability $p_\btheta(\bsx\mis_{1:T}|\bsz_{1:T})$ would guide the model to learn to correctly impute those missing values. Without such true values, we can only encourage the model to impute
the missing values with some indirect criteria. In the objective \eqref{eq:supnotmiwae_bound}, there are two terms that the model hinges on for this matter.

\begin{itemize}[leftmargin=7mm]
    \item The missing model $p_\bpsi(\bss_{1:T}|\bsx_{1:T}\obs, \bsx_{1:T}\mis)$:
this term encourages the model to reconstruct the missing mask $s_t$ from the imputed value $x_t\mis$, so in principle, the model should impute the missing values in a way that they are distinguishable from the observed values. However, in general, the distributions of the observed and the missings are not necessarily different, and more importantly, the model can easily cheat the objective. For instance, consider a trivial case where the model imputes all the missing values with zero. The conditional probability $p_\bpsi(\bss_{1:T}|\bsx_{1:T}\obs, \bsx_{1:T}\mis)$ can still be maximized by setting $\sigma_\text{mis}(x_{t,j}) = 0$ if $x_{t,j} = 0$ (unless there are not many observed with $x\obs_{t,j}=0$).
    \item The classifier $p_\btheta(\bsy|\bsx_{1:T}\obs, \bsx\mis_{1:T})$: this term expects the model to generate meaningful imputations so that they are helpful for the classification. However, as shown in prior works~\citep{che2018recurrent}, the classifier can achieve decent classification accuracy \emph{without} meaningful imputations, for instance, it will still be able to classify the signals while all the missing values are imputed with zeros. Hence, in the current form, there is no strong incentive for the model to learn non-trivial imputations that will bring significant accuracy gain over zero imputations.
\end{itemize}

To summarize, a model trained with the objective \eqref{eq:supnotmiwae_bound} is not likely to generate realistic missing values. To resolve this, we may introduce a missing model $p_\btheta(\bss_{1:T}|\bsx\obs_{1:T}, \bsx\mis_{1:T})$ much more elaborated than the simple i.i.d. model that we are using right now, but that may require some dataset-specific design. Instead, we present a simple regularization technique that can effectively enhance the quality of the imputed values.

Our idea is simple; when passing the observed inputs $\bsx\obs_{1:T}$ and the imputed missing values $\hat{\bsx}\mis_{1:T}$ (i.e., imputed by \eqref{eq:decayed_impute}) to the classifier, \emph{deliberately drop} some portion of the observed inputs. Without dropping the observed inputs, the classifier may heavily rely on the observed inputs to do the classification, but if some of the observed inputs are dropped out during training, the classifier can focus more on the imputed missing values $\hat{\bsx}_{1:T}\mis$. As a result, the model is encouraged to generate more ``useful" missing values that are beneficial for classification. More specifically, let $\beta$ be a predefined dropout probability. Then we construct the imputed input $\hat{\bsx}_t$ to the classifier as follows:
\[
\tilde{\bsx}_{1:T} &:= (\bsx_{1:T}\obs, \bsx_{1:T}\mis) \text{ where } \bsx_{1:T}\mis \sim p_\btheta(\bsx_{1:T}\mis|\bsz_{1:T}),\nonumber\\
m_{t,j} &\sim \mathrm{Bern}(1-\beta),\nonumber\\
\hat{x}_{t,j} &:= \mathrm{Decay}(s_{t,j}m_{t,j},  x_{t,j},\gamma_{\hat{\bsx}_t,j}, x_{t',j}, \tilde{x}_{t,j}).
\]
That is, when an observed $x_{t,j}$ is dropped out, we put a generated value with the decay applied as in \eqref{eq:decayed_impute}, so that the classifier could focus more on the values generated by the decoder as we intended. We call this idea \emph{ObsDropout}, since we are dropping out the observed values during the training. 

With the mask variables $\bsm_{1:T}$ included, the likelihood is extended
\[
\lefteqn{p_{\btheta,\bpsi, \blambda}(\bsy, \bsx_{1:T}\obs, \bss_{1:T})}\nonumber\\
&=\int p_\blambda(\bsy|\bsx\obs_{1:T}, \bsx\mis_{1:T}, \bsm_{1:T}) p_\beta(\bsm_{1:T})\nonumber\\
& \quad\times p_\btheta(\bsx\obs_{1:T}|\bsz_{1:T})p_\btheta(\bsx\mis_{1:T}|\bsz_{1:T})\nonumber\\
& \quad\times p_\btheta(\bz_{1:T})
p_\bpsi(\bss_{1:T}|\bsx_{1:T}) \dee \bsx\mis_{1:T} \dee \bsz_{1:T} \dee\bsm_{1:T}.
\]
The corresponding IWAE objective is defined similarly to \cref{eq:supnotmiwae_bound}, with the expectation taken with respect to $K$ copies of a variational distribution, $\prod_{k=1}^K p_\btheta(\bsx_{k,1:T}\mis|\bsz_{k,1:T}) q_\bphi(\bsz_{k,1:T}|\bsx_{1:T}\obs)p_\beta(\bsm_{k,1:T})$ and the importance term is defined as
\[
\omega_k &:= p_\btheta(\bsy|\bsx_{1:T}\obs, \bsx\mis_{k,1:T}, \bsm_{k,1:T}) p_\bpsi(\bss_{1:T}|\bsx_{1:T}\obs, \bsx\mis_{k,1:T}) \nonumber\\
& \times p_\btheta(\bsx_{1:T}\obs|\bsz_{k,1:T}) p_\btheta(\bsz_{k,1:T}) / q_\bphi(\bsz_{k,1:T}|\bsx_{1:T}\obs),
\]
where $p_\btheta(\bsm_{1:T}) := \prod_{t=1}^T \prod_{j=1}^d \mathrm{Bern}(m_{t,j}|\beta)$.

\subsection{Prediction}
Similar to SupMIWAE, we exploit Self-Normalized Importance Sampling (SNIS) to approximate the predictive distribution for a new input $\bsx_{1:T}\obs$. With the model trained with obsdropout, we have
\[\label{eq:pred_drop}
p(\bsy|\bsx\obs_{1:T})
\approx \frac{1}{S}\sum_{s=1}^S\sum_{k=1}^K \bar{\zeta}^{(s)}_k,
\]
where
\[
\lefteqn{(\bsz^{(s)}_{k,1:T}, (\bsx\mis)^{(s)}_{k,1:T}, \bsm^{(s)}_{k,1:T})}\nonumber\\
&\iidsim q_\bphi(\bsz_{1:T}|\bsx\obs_{1:T}) p_\btheta(\bsx\mis_{1:T}|\bsz_{1:T})
p_\beta(\bsm_{1:T}),\\
\zeta_k^{(s)} &:= p_\blambda(\bsy| \bsx\obs_{1:T}, (\bsx\mis)^{(s)}_{k,1:T}, \bsm^{(s)}_{k,1:T}) p_\btheta(\bsx\obs_{1:T}|\bsz^{(s)}_{k,1:T})\nonumber\\
&\quad \times p_\btheta(\bsz^{(s)}_{k,1:T})/q_\bphi(\bsz^{(s)}_{k,1:T}|\bsx\obs_{1:T}),\\
\bar\zeta^{(s)}_k &:= \zeta_k^{(s)} / \sum_{\ell=1}^k \zeta^{(s)}_\ell.
\]

% Note that here we are still dropping out the observations. We can instead choose not to apply ObsDropout during the prediction, and the corresponding predictive distribution is approximated as,
% \[\label{eq:pred_no_drop}
% p(\bsy|\bsx\obs_{1:T}) \approx \frac{1}{L}\sum_{s=1}^S\sum_{k=1}^K \bar{\omega}^{(s)}_k p_\blambda(\bsy| \bsx\obs_{1:T}, (\bsx\mis)^{(s)}_{k,1:T}),
% \]
% without sampling $\bsm_{k,1:T}^{(s)}$. We compare both of this predictive distributions in the experiments.

\section{Related Work}
\label{main:sec:related}
There are two lines of literature closely related to our method. The first line consists of work dealing with the problem of missing data imputation based on Deep Latent Variable Models (DLVMs). The other line consists of work designing a tailored neural network architecture for supervised learning on irregularly sampled time series data, which also implies missingness in the process of handling it as a tensor.

\paragraph{DLVMs for missing data} \citet{mattei2019miwae} proposed the MIWAE bound for training DLVMs in the presence of missing data under the MAR assumption. \citet{ipsen2021notmiwae} modified the MIWAE bound suitable for the MNAR scenario. \citet{ipsen2022how} extended the MIWAE bound to the supervised learning task. This line of work provides a useful framework for training DLVMs under missingness. However, it is not directly applicable to time series data because it cannot model the temporal dependency within a series. There is previous work that makes DLVMs suitable for multivariate time series. For example, \citet{forutin2020gpvae} proposed a CNN-based VAE architecture with a Gaussian Process prior to encode the temporal correlation in the latent space. \citet{rubanova2019ode} presented ODE-RNNs, which employ Neural Ordinary Differential Equations (Neural ODEs) to model hidden state dynamics of RNNs. \citet{shukla2021heteroscedastic} developed an attention-based VAE architecture with probabilistic interpolation for irregularly sampled time series data.

\paragraph{Irregularly sampled time series classification} Researchers have developed deep neural network architectures customized to classify irregularly sampled time series data. Several architectures have shown competitive empirical performance in this task. \citet{che2018recurrent} modified the architecture of GRU intending to perform supervised learning with sparse covariates by introducing a learnable temporal decay mechanism for the input and hidden state of GRU. This mechanism has been applied to further research. For example, \citet{cao2018brits} employed temporal decay in hidden states of their bidirectional-RNN-based model to capture the missing pattern of irregularly sampled times series. \citet{shukla2019interpnet} presented a hybrid architecture of an interpolation network and a classifier. The interpolation network takes irregularly sampled time series as input and returns fully observed and regularly sampled representation of the original time series data. \citet{shukla2021multi} later modified interpolation network with an attention-based architecture.
\section{Experiments}
\label{main:sec:experiments}

In this section, we demonstrate our method on real-world multivariate time series data with missing values. We compare ours to the baselines on three datasets: MIMIC-III~\citep{mimic3}, PhysioNet 2012~\citep{physionet2012}, and Human Activity Recognition~\citep{humanactivity}. MIMIC-III and PhysioNet 2012 datasets contain Electronic Health Records of patients from Intensive Care Units (ICU). Human Activity Recognition dataset consists of the 3D coordinate of sensors mounted on the people doing some daily activities such as walking and sitting. See \cref{app:sec:details} for the details of datasets. 
For all three datasets, we compare classification accuracy and the uncertainty quantification performances in \cref{main:subsec:classification}.
We also compare the missing value imputation performance of our methods to the baselines in \cref{main:subsec:imputation}

For the baselines, we considered GRU classifiers with various imputation methods and other deep neural network based methods that are considered to be competitive in the literature. See \cref{app:sec:details} for the detailed description of the baselines. For the uncertainty quantification metrics, we compared cross-entropy (CE, equals negative log-likelihood), expected calibration error (ECE), and brier score (BS).
Please refer to \cref{app:sec:details} for the detailed description of the metrics.

\subsection{Classification Results}
\label{main:subsec:classification}

\begin{table}[t]
    \caption{Classification performances of baseline methods and ours on MIMIC-III dataset.}
    \centering
    \def\arraystretch{1.1}
    \vskip 0.05in
    \resizebox{\columnwidth}{!}{
        \begin{tabular}{lllll}
            \toprule
            Method               & \umet{AUROC}         & \dmet{CE}            & \dmet{ECE}           & \dmet{BS}            \\
            \midrule
            GRU-Mean             & \tv[n]{0.844}{0.001} & \tv[n]{0.477}{0.012} & \tv[n]{0.240}{0.010} & \tv[n]{0.154}{0.004} \\
            GRU-Simple           & \tv[n]{0.824}{0.003} & \tv[n]{0.451}{0.053} & \tv[n]{0.197}{0.039} & \tv[g]{0.150}{0.020} \\
            GRU-Forward          & \tv[n]{0.856}{0.002} & \tv[n]{0.455}{0.014} & \tv[n]{0.219}{0.012} & \tv[n]{0.149}{0.005} \\
            GRU-D                & \tv[n]{0.855}{0.001} & \tv[n]{0.468}{0.011} & \tv[n]{0.224}{0.009} & \tv[n]{0.157}{0.004} \\
            PhasedLSTM           & \tv[n]{0.802}{0.005} & \tv[n]{0.499}{0.049} & \tv[n]{0.229}{0.033} & \tv[n]{0.166}{0.019} \\
            IP-Nets              & \tv[n]{0.830}{0.005} & \tv[n]{0.509}{0.031} & \tv[n]{0.253}{0.020} & \tv[n]{0.172}{0.012} \\
            SeFT                 & \tv[n]{0.837}{0.003} & \tv[b]{0.432}{0.013} & \tv[b]{0.186}{0.008} & \tv[n]{0.142}{0.005} \\
            \cmidrule(lr){1-5}
            Ours                 & \tv[b]{0.860}{0.003} & \tv[b]{0.432}{0.016} & \tv[n]{0.197}{0.011} & \tv[b]{0.141}{0.008} \\
            \quad w/o obsdropout & \tv[n]{0.849}{0.003} & \tv[n]{0.490}{0.024} & \tv[n]{0.232}{0.015} & \tv[n]{0.161}{0.009} \\
            \quad w/o MNAR       & \tv[n]{0.851}{0.001} & \tv[n]{0.462}{0.007} & \tv[n]{0.220}{0.007} & \tv[n]{0.151}{0.003} \\
            \bottomrule
        \end{tabular}
    }
    \label{tab:m3_classification}
\end{table}

\begin{table}[t]
    \caption{Classification performances of baseline methods and ours on PhysioNet 2012 dataset.}
    \centering
    \def\arraystretch{1.1}
    \vskip 0.05in
    \resizebox{\columnwidth}{!}{
        \begin{tabular}{lllll}
            \toprule
            Method               & \umet{AUROC}         & \dmet{CE}            & \dmet{ECE}           & \dmet{BS}            \\
            \midrule
            GRU-Mean             & \tv[n]{0.853}{0.003} & \tv[n]{0.439}{0.013} & \tv[n]{0.194}{0.013} & \tv[n]{0.144}{0.006} \\
            GRU-Simple           & \tv[n]{0.819}{0.005} & \tv[n]{0.457}{0.065} & \tv[n]{0.145}{0.058} & \tv[n]{0.150}{0.026} \\
            GRU-Forward          & \tv[n]{0.850}{0.006} & \tv[n]{0.475}{0.028} & \tv[n]{0.221}{0.026} & \tv[n]{0.156}{0.010} \\
            GRU-D                & \tv[n]{0.862}{0.003} & \tv[n]{0.416}{0.024} & \tv[n]{0.166}{0.025} & \tv[n]{0.135}{0.009} \\
            PhasedLSTM           & \tv[n]{0.802}{0.007} & \tv[n]{0.489}{0.089} & \tv[n]{0.199}{0.074} & \tv[n]{0.166}{0.035} \\
            IP-Nets              & \tv[n]{0.863}{0.003} & \tv[n]{0.424}{0.041} & \tv[n]{0.180}{0.043} & \tv[n]{0.138}{0.015} \\
            SeFT                 & \tv[n]{0.863}{0.001} & \tv[n]{0.458}{0.044} & \tv[n]{0.209}{0.039} & \tv[n]{0.156}{0.015} \\
            \cmidrule(lr){1-5}
            Ours                 & \tv[b]{0.874}{0.004} & \tv[n]{0.399}{0.057} & \tv[n]{0.154}{0.048} & \tv[n]{0.129}{0.021} \\
            \quad w/o obsdropout & \tv[n]{0.866}{0.004} & \tv[b]{0.371}{0.050} & \tv[b]{0.119}{0.041} & \tv[b]{0.116}{0.018} \\
            \quad w/o MNAR       & \tv[n]{0.869}{0.001} & \tv[n]{0.429}{0.033} & \tv[n]{0.182}{0.019} & \tv[n]{0.140}{0.012} \\
            \bottomrule
        \end{tabular}
    }
    \label{tab:p12_classification}
\end{table}

\begin{table}[t]
    \centering
    \caption{Classification performance of baseline methods and ours on Human Activity Recognition dataset.}
    \def\arraystretch{1.1}
    \vskip 0.05in
    \resizebox{\columnwidth}{!}{%
        \begin{tabular}{lllll}
            \toprule
            Method       & \umet{Accuracy}      & \dmet{CE}            & \dmet{ECE}           & \dmet{BS}            \\
            \midrule
            GRU-Mean     & \tv[n]{0.780}{0.005} & \tv[n]{0.163}{0.012} & \tv[n]{0.019}{0.006} & \tv[n]{0.046}{0.002} \\ 
            GRU-Simple   & \tv[n]{0.861}{0.005} & \tv[n]{0.070}{0.004} & \tv[n]{0.010}{0.002} & \tv[n]{0.019}{0.001} \\
            GRU-Forward  & \tv[n]{0.847}{0.006} & \tv[n]{0.084}{0.008} & \tv[n]{0.012}{0.003} & \tv[n]{0.021}{0.001} \\ 
            GRU-D        & \tv[n]{0.860}{0.005} & \tv[n]{0.081}{0.005} & \tv[n]{0.012}{0.001} & \tv[n]{0.020}{0.001} \\ 
            PhasedLSTM   & \tv[n]{0.852}{0.001} & \tv[n]{0.070}{0.004} & \tv[n]{0.008}{0.003} & \tv[n]{0.020}{0.001} \\
            SeFT    & \tv[n]{0.848}{0.005} & \tv[n]{0.068}{0.001} & \tv[b]{0.003}{0.001} & \tv[n]{0.020}{0.001} \\
            \cmidrule(lr){1-5}
            Ours                 & \tv[b]{0.883}{0.003} & \tv[b]{0.063}{0.003} & \tv[n]{0.009}{0.001} & \tv[b]{0.016}{0.000} \\
            \quad w/o obsdropout & \tv[n]{0.867}{0.018} & \tv[n]{0.068}{0.004} & \tv[n]{0.008}{0.003} & \tv[n]{0.018}{0.000} \\
            \quad w/o MNAR & \tv[n]{0.882}{0.004} & \tv[n]{0.070}{0.004} & \tv[n]{0.011}{0.001} & \tv[n]{0.017}{0.000} \\
            \bottomrule
        \end{tabular}
    }
    \label{tab:uha_classification}
\end{table}

We summarize the classification results in \cref{tab:m3_classification}, \cref{tab:p12_classification}, and \cref{tab:uha_classification}.
In general, our method achieves the best performance among the competing methods both in terms of prediction accuracy and uncertainty quantification. In all classification experiments, our method beats other baseline methods by a wide margin in terms of predictive accuracy. Also, our model shows competitive results with respect to uncertainty metrics. Although SeFT~\citep{horn2020set} shows strong results in terms of uncertainty quantification in some experiments, this method shows inferior performance with respect to predictive accuracy. Since it is reasonable to compare uncertainty quantification between models with similar predictive performance, it can be said that our model shows the best performance among baseline models in general. We also provide an ablation study for our model to see the effect of 1) obsdropout and 2) MNAR assumption. The results clearly show that both components play important roles in our model. For all the experiments, obsdropout clearly makes the gain in terms of predictive performance. Also, removing MNAR assumption makes the worse performance.

% In \cref{app:sec:additional}, we provide further results showing the effect of dropout rate $\beta$ for the performance.

% \begin{figure*}[t]
%      \centering
%      \includegraphics[width=0.3\textwidth]{figure/mlpdecoder_18.png}
%      \includegraphics[width=0.3\textwidth]{figure/grudecoder_18.png}
%      \caption{Plots of $\bmu_{\text{dec}}(\bsz_{1:t}), \bsigma^2_{\text{dec}}(\bsz_{1:t})$ where the encoder and the decoder are MLPs. This generates unrealistic spike since model does not take temporal correlation into account}
%      \label{fig:decoder_imputation}
%  \end{figure*}

\begin{figure*}[t]
    \vspace*{2mm}
     \centering
     \includegraphics[width=0.45\textwidth]{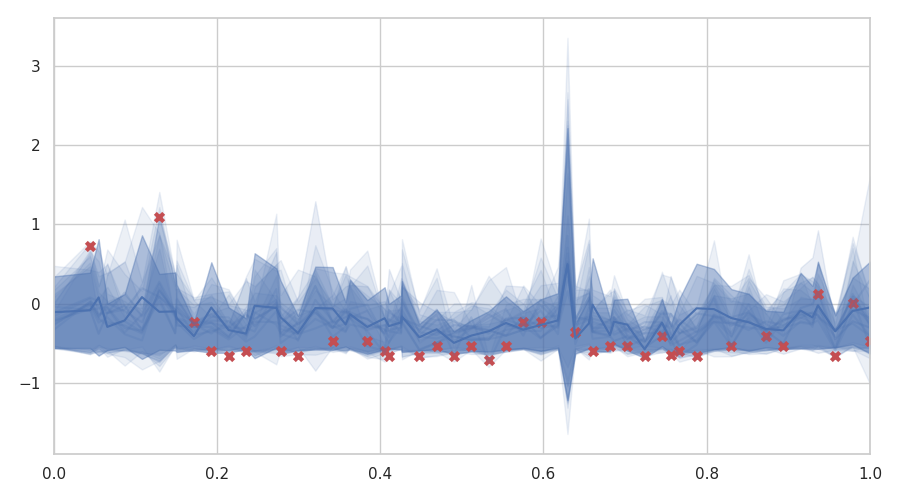}
     \includegraphics[width=0.45\textwidth]{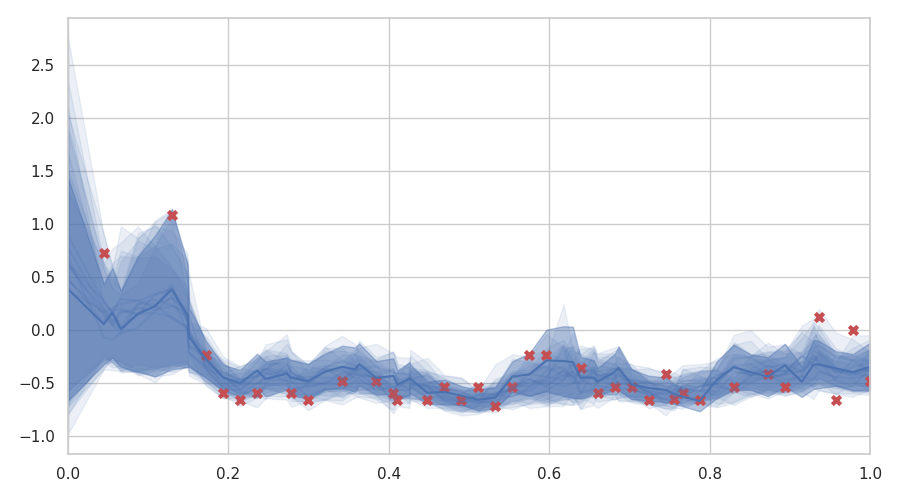}
     \caption{Plots of $\bmu_{\text{dec}}(\bsz_{1:t}), \bsigma^2_{\text{dec}}(\bsz_{1:t})$. (\textbf{Left}) Our model with MLP encoder and MLP decoder. (\textbf{Right}) Our model trained with obsdropout with a rate of 0.4. Since MLP architecture does not take the temporal association into account, it shows spiky imputation while our model shows smooth imputation. Also, our model shows better performance in uncertainty quantification since the learned variance of the decoder captures sudden spikes and considers the initial part of the time series more uncertain.}
     \label{fig:dropout_imputation}
 \end{figure*}
 
\subsection{Imputation Results}
\label{main:subsec:imputation}

\begin{table*}[t]
    \centering
    \caption{Imputation performance on PhysioNet 2012, MIMIC-III and Human Activity Recognition dataset.}
    \def\arraystretch{1.1}
    \vskip 0.05in
    \resizebox{0.9\textwidth}{!}{%
        \begin{tabular}{lllllll}
            \toprule
                                                     & \multicolumn{2}{c}{PhysioNet 2012}          & \multicolumn{2}{c}{MIMIC-III}               & \multicolumn{2}{c}{Human Activity Recognition} \\
                                                       \cmidrule(lr){2-3}                            \cmidrule(lr){4-5}                            \cmidrule(lr){6-7}
            Method                                   & \dmet{MAE}           & \dmet{MRE}           & \dmet{MAE}           & \dmet{MRE}           & \dmet{MAE}           & \dmet{MRE}           \\
            \midrule
            Mean                                     & \tv[n]{0.696}{0.001} & \tv[n]{1.000}{0.000} & \tv[n]{0.330}{0.002} & \tv[n]{1.000}{0.000} & \tv[n]{0.799}{0.001} & \tv[n]{1.000}{0.000} \\
            Forward                                  & \tv[n]{0.400}{0.004} & \tv[n]{0.576}{0.005} & \tv[n]{0.151}{0.002} & \tv[n]{0.459}{0.002} & \tv[n]{0.305}{0.005} & \tv[b]{0.373}{0.007} \\
            GP-VAE                                   & \tv[n]{0.439}{0.010} & \tv[n]{0.630}{0.005} & \tv[n]{0.198}{0.012} & \tv[n]{0.601}{0.031} & \tv[n]{0.548}{0.008} & \tv[n]{0.684}{0.019} \\
            SAITS                                    & \tv[n]{0.653}{0.030} & \tv[n]{0.942}{0.013} & \tv[n]{0.341}{0.005} & \tv[n]{1.040}{0.087} & \tv[n]{0.834}{0.005} & \tv[n]{1.048}{0.013} \\
            \cmidrule(lr){1-7}
            Ours                                     & \tv[b]{0.367}{0.005} & \tv[b]{0.526}{0.005} & \tv[n]{0.149}{0.002} & \tv[n]{0.451}{0.008} & \tv[b]{0.297}{0.005} & \tv[b]{0.373}{0.006} \\
            \qquad w/o supervision                   & \tv[n]{0.376}{0.007} & \tv[n]{0.541}{0.009} & \tv[b]{0.148}{0.002} & \tv[b]{0.449}{0.006} & \tv[n]{0.298}{0.005} & \tv[b]{0.373}{0.007} \\
            \qquad w/o obsdropout                    & \tv[n]{0.377}{0.004} & \tv[n]{0.542}{0.005} & \tv[n]{0.152}{0.001} & \tv[n]{0.459}{0.002} & \tv[n]{0.299}{0.005} & \tv[n]{0.374}{0.006} \\
            \qquad w/o supervision \& MNAR        & \tv[n]{0.394}{0.003} & \tv[n]{0.570}{0.006} & \tv[n]{0.150}{0.002} & \tv[n]{0.457}{0.003} & \tv[n]{0.299}{0.005} & \tv[n]{0.374}{0.006}
            \\
            \bottomrule
        \end{tabular}
    }
    \label{tab:imputation}
\end{table*}

We quantitatively check the imputation performance of our model on three datasets in \cref{tab:imputation}, and visually check the imputation quality by changing our model settings in \cref{fig:dropout_imputation}. Although our model is designed for the classification, ours achieved the lowest MAE and MRE, outperforming the baseline (SAITS) specifically designed for the imputation. Also, our model achieved better performance than GP-VAE model which employs identical prior distribution. This implies that our architecture design and other ingredients such as obsdropout and missing model and supervised signal helps the imputation. Only forward imputation shows comparable performance with our model. However, the structure of forward imputation is very restricted so it cannot deal with various time series such as time series that have many sudden spikes or periodicity. Since our model simultaneously employs the decay mechanism and generative model, our model is more flexible and able to cope with various cases. Especially, the ablation study on the class supervision part $p_\blambda(\bsy|\bsx_{1:T})$ and the obsdropout and MNAR assumption implies that the imputation values generated by our model which was trained to better classify the signals are more ``realistic". \cref{fig:dropout_imputation} highlights the effect of using transformer-based encoders and decoders. The values imputed with those techniques form smoother trajectories and better capture the uncertainties in the intervals without observed values.

\section{Conclusion}
\label{main:sec:conclusion}

In this paper, we presented a novel probabilistic framework for multivariate time series classification with missing data. Under the MNAR assumption, we first developed a deep generative model suitable for generating missing values in multivariate time series data. Then we identified an important drawback of the na\"ive combination of the deep generative models with the classifiers and proposed a novel regularization technique called obsdropout to circumvent that.
In this way, combining the MNAR assumption and the obsdropout regularization technique, the generative model can generate more natural imputation, and the classifier can also perform more accurate and robust classification through this.
Also, by using the transformer layers in the internal modules, it can effectively capture the time series structure.
Through experiments, we show that it is possible to achieve high performance and uncertainty calibration at the same time in classification tasks with missing values.
We demonstrated that ours could classify real-world multivariate time series data more accurately and robustly than existing methods.
% In this paper, we focused on transformer-based architectures for both generative models and classifiers.
% An interesting future work would be extending our methods with other architectures such as transformers~. 
% In this paper, we focus on transformer-based architectures for both generative models and classifiers because transformers can effectively capture the structure of a time series.

\paragraph{Reproducibility statement}
Please refer to \cref{app:sec:details} for full experimental detail including datasets, models, and evaluation metrics.

\section*{Acknowledgement}
This work was partially supported by Institute of Information \& communications Technology Planning \& Evaluation (IITP) grant funded by the Korea government (MSIT) (No.2019-0-00075, Artificial Intelligence Graduate School Program (KAIST)), Artificial Intelligence Innovation Hub (No.2022-0-00713), and National Research Foundation of Korea (NRF) funded by the Ministry of Education (NRF2021M3E5D9025030).

\bibliography{references}
\bibliographystyle{icml2023}

%%%%%%%%%%%%%%%%%%%%%%%%%%%%%%%%%%%%%%%%%%%%%%%%%%%%%%%%%%%%%%%%%%%%%%%%%%%%%%%
%%%%%%%%%%%%%%%%%%%%%%%%%%%%%%%%%%%%%%%%%%%%%%%%%%%%%%%%%%%%%%%%%%%%%%%%%%%%%%%
% APPENDIX
%%%%%%%%%%%%%%%%%%%%%%%%%%%%%%%%%%%%%%%%%%%%%%%%%%%%%%%%%%%%%%%%%%%%%%%%%%%%%%%
%%%%%%%%%%%%%%%%%%%%%%%%%%%%%%%%%%%%%%%%%%%%%%%%%%%%%%%%%%%%%%%%%%%%%%%%%%%%%%%
\newpage
\appendix
\onecolumn
\section{Experimental Details}
\label{app:sec:details}

\subsection{Datasets}

\subsubsection{Dataset Description}
We use three irregularly sampled time series datasets to evaluate the classification and imputation performance of our model and baseline models.

\paragraph{PhysioNet Challenge 2012 (PhysioNet 2012)}
This dataset contains approximately 12,000 Electronic Health Records of adult patients who were admitted to the intensive care unit (ICU). Each record contains up to 37 time series variables including vital signs such as heart rate, and temperature. All of the variables were measured during the first 48 hours of each patient's admission to ICU and the sampling rate of times series varies among variables. After preprocessing, we have 37 features and 11,971 data points. This data is extremely sparse, with 85\% of the entries being unobserved.

\paragraph{MIMIC-III}
MIMIC-III dataset is a widely used database that includes de-identified Electronical Health Record of patients who stayed in ICU of Beth Israel Deaconess Medical Center from 2001 to 2012. It originally consists of approximately 57,000 records of patients who stayed in ICU. Records include various variables such as medications, in-hospital mortality, and vital signs. ~\citet{harutyunyan2019multitask} set variety of benchmark tasks based on the subset of this database. Among them, we conduct binary in-hospital mortality prediction task for measuring classification performance.
After preprocessing, our dataset contains 16 features and 21,107 data points. 

\paragraph{UCI Localization Data for Person Activity (UCI Human Activity)}
This dataset includes records of five people doing some usual activities such as walking or sitting. All people wear sensors on their right ankle, left, belt and chest. During activities, the sensors record their position in the form of three-dimensional coordinates at very short intervals. Activities of each people at a certain time point are classified into one of 11 classes and recorded with the position of sensors. After preprocessing, we have total 6554 time series with 12 features (3-dimensional coordinates of 4 devices). Using this data, we conduct an online-prediction task. The objective of this task is to classify each individual's activity per time point based on the position of sensors.

\begin{table}[t]
    \caption{Statistics of each dataset.}
    \centering
    \vskip 0.05in
    \resizebox{0.6\textwidth}{!}{
        \begin{tabular}{lrrr}
        \toprule
                                     & PhysioNet2012 & MIMIC-III & Human Activity \\
        \midrule
        Number of samples            &        11,971 &    21,107 &           6554 \\
        Number of variables          &            37 &        16 &             12 \\
        Maximum number of time steps &           215 &       292 &            50 \\
        Mean of number of time steps &            74 &        78 &          162.4 \\
        Total missing rate           &         0.843 &     0.655 &          0.75 \\
        \bottomrule
        \end{tabular}
    }
    \label{tab:dataset_statistics}
\end{table}

\subsubsection{Data preprocessing}
For all datasets, we basically standardize the numerical covariates so that all features have zero mean and unit variance, respectively. Also, we normalize the time scale to be in $[0,1]$ scale.
\paragraph{PhysioNet2012 and MIMIC-III}

Since there is no fixed rule for preprocessing PhysioNet 2012 dataset and MIMIC-III database researchers usually preprocess the raw data on their own. Therefore, it is difficult for practitioners to compare experimental results with other works. So, for comparability, we employ python package \texttt{medical\_ ts\_datasets}\footnote{\href{https://github.com/ExpectationMax/medical_ts_datasets}{https://github.com/ExpectationMax/medical\_ts\_datasets}} ~\citep{horn2020set} which provides the unified data preprocessing pipeline for Physionet2012 and MIMIC-III datasets. For both datasets, patients who have more than 1000 time steps or have no observed time series data were excluded. Also, discretizing the time step of data by an hour and aggregating the measurement is frequently used to preprocess Physionet2012 in previous work~\citep{rubanova2019ode}, but this package preserves much more original time series variables while preprocessing than hourly-based aggregation preprocessing for both datasets. 
Please refer to ~\citep{horn2020set} for more details of dataset preprocessing.

\paragraph{UCI Human Activity} 

Basically, we decide to follow the preprocessing of this dataset based on ~\citep{rubanova2019ode}, except for aggregation of time points of all data points in the dataset. We just use overlapping intervals of 50 time points while ~\citep{rubanova2019ode} used 221 time points. So, our processed data finally contains 6554 data points and 50 time points, and 12 features.

\subsection{Details for classification experiments} %Experimental details}
For all experiments, we use five different random seeds to conduct experiments.
Then we measure the mean and the variance of each value.

\subsubsection{baseline methods}
\begin{itemize}
    \item \textbf{GRU-mean}: Missing value is simply replaced with the empirical mean of each variable.
    \item \textbf{GRU-forward}: Missing entries are filled with previously observed values.
    \item \textbf{GRU-simple}: concatenate the mask $\bss_t$, and the \emph{time-interval} $\bdelta_t$ along with the imputed vector $\hat{\bsx}_t$. The concatenated vector $[\hat{\bsx}_t, \bss_t, \bdelta_t]$ is then fed into GRU.
    \item \textbf{GRU-D}: Missing values are imputed as a weighted mean of the last observed $x_{t',j}$ and the mean $\bar{x}_j$ with the learnable weight.
     \item \textbf{Phased-LSTM}: This model is LSTM variant designed to deal with long sequence input by introducing a time gate in their cell to prevent memory decay when useful information is absent for a long time~\citep{neil2016phased}.
     \item \textbf{Interpolation-Prediction Network (IP-Nets)}: Instead of directly imputing missing values, IP-Nets \citep{shukla2019interpnet} employed semi-parametric interpolation network that makes regularly spaced representation of irregularly sampled time series data. Then, this representation is fed into a prediction network such as GRU. 
     \item \textbf{Set Function for Time Series (SeFT)}: SeFT \citep{horn2020set} is a set function-based method attached with an attention mechanism for multivariate time series classification task. This model encodes the observed value of the time series as a set and uses the attention layer to aggregate the embedding of elements. Since SeFT only uses the observed value as input, imputation is unnecessary.
\end{itemize}

\subsubsection{Model implementation in Human Activity experiment}

Since we conduct online-prediction task on Human Activity dataset, we do not consider IP-Nets as baseline models because this model exploits future information when conducting interpolation before feeds the representation to the classification model. SeFT also use future information due to the set encoding, but the author offers the method to prevent information leakage by calculating attention weights in cumulative manner. Our method use causal masked attention layer in encoder and decoder, we do not use future information during generation. For classifier, we use causal mask for attention layer in classifier for this experiment.

\subsubsection{Training details}

% For all the classification  Also, we use Adam optimizer with learning rate 0.0001 and batch size 128 for all models for MIMIC-III in-hospital mortality prediction task. In PhysioNet 2012 experiment, we use Adam optimizer with learning rate 0.001 and batch size 128 for all models. For online-prediction task, we also employ Adam optimizer with learning rate 0.001 and batch size 128.

For all the classification experiments, we fix a batch size of 128. We adopt Adam with weight decay as the optimizer and find the best weight decay for each model using grid search. 

We conduct a grid search to find the best hyperparameters of each model. See the table for concrete search space and the best hyperparameters for each model.

We employ early stopping for all classification experiments. We set early stopping patience to 20 epochs and set the valid Area Under ROC curve(AUROC) as the early stopping criterion. Since the label imbalance of PhysioNet 2012 and MIMIC-III is extreme, we over-sample the mortality class to train models on the balanced batches. 

\subsubsection{Hyperparameters}

We search all hyperparameters in the grid to find the best hyperparameters for each model.
For all models, we search the weight decay of the AdamW optimizer in $\{0, 0.01, 0.1, 1\}$ and the number of units (n\_units) of model layers in $\{128, 256\}$ if the model accepts that parameter.
For GRU variant models, we search the dropout probability of forward pass (dropout) in $\{0, 0.1, 0.2, 0.3\}$, and the dropout probability of recurrent pass (recurrent\_dropout) in $\{0, 0.1, 0.2, 0.3\}$.
For the SeFT-Attn model, we followed the best hyperparameter settings presented in ~\citet{horn2020set}.
See Table~\ref{tab:hyperparam_classification} for hyperparameter settings of our model and baseline methods for all classification experiments.

\begin{table*}[h]
    \newcommand{\crule}{\cmidrule(lr){2-3}}
    \caption{Hyperparameter settings for classification experiments}
    \centering
    \vskip 0.05in
    \def\arraystretch{1.1}
    \resizebox{0.95\textwidth}{!}{
        \begin{tabular}{lll}
            \toprule
            Dataset  & Method  & Hyperparameters \\
            \midrule
            \multirow{12}{*}{PhysioNet2012}
            & GRU-Mean              & weight\_decay: 0.1, n\_units: 128, dropout: 0.3, recurrent\_dropout: 0.1 \\
            & GRU-Simple            & weight\_decay: 1.0, n\_units: 128, dropout: 0.0, recurrent\_dropout: 0.1 \\
            & GRU-Forward           & weight\_decay: 0.0, n\_units: 256, dropout: 0.0, recurrent\_dropout: 0.2 \\
            & GRU-D                 & weight\_decay: 1.0, n\_units: 128, dropout: 0.0, recurrent\_dropout: 0.2 \\
            \crule
            & PhasedLSTM            & weight\_decay: 0.0, n\_units: 256, use\_peepholes: False, leak: 0.01, period\_init\_max: 1000.0 \\
            & IP-Nets               & weight\_decay: 1.0, n\_units: 128, imputation\_stepsize: 1 , reconst\_fraction: 0.5 \\
            & Seft                  & weight\_decay: 0.0, n\_phi\_layers: 4, phi\_width: 128, phi\_dropout: 0.2, n\_psi\_layers: 2 \\
            &                       & psi\_width: 64, psi\_latent\_width: 128, dot\_prod\_dim: 128, n\_heads: 4 \\
            &                       & attn\_dropout: 0.5, latent\_width: 32, n\_rho\_layers: 2, rho\_width: 512 \\
            &                       & rho\_dropout: 0.0, max\_timescale: 100., n\_positional\_dims: 4 \\
            \crule
            & Ours                  & weight\_decay: 0.0, n\_train\_latents: 10, n\_train\_samples: 1, n\_test\_latents: 20, n\_test\_samples: 30, \\  %TODO
            &                       & n\_hidden: 128, z\_dim: 32, n\_units: 128, observe\_dropout: 0.4 \\
            \midrule
            \multirow{12}{*}{MIMIC-III}
            & GRU-Mean              & weight\_decay: 1.0, n\_units: 128, dropout: 0.1, recurrent\_dropout: 0.2 \\
            & GRU-Simple            & weight\_decay: 1.0, n\_units: 256, dropout: 0.0, recurrent\_dropout: 0.1 \\
            & GRU-Forward           & weight\_decay: 0.0, n\_units: 256, dropout: 0.0, recurrent\_dropout: 0.1 \\
            & GRU-D                 & weight\_decay: 1.0, n\_units: 128, dropout: 0.0, recurrent\_dropout: 0.2 \\
            \crule
            & PhasedLSTM            & weight\_decay: 1.0, n\_units: 128, use\_peepholes: False, leak: 0.01, period\_init\_max: 1000.0 \\
            & IP-Nets               & weight\_decay: 1.0, n\_units: 256, imputation\_stepsize: 1, reconst\_fraction: 0.5 \\
            & Seft                  & weight\_decay: 0.0, n\_phi\_layers: 3, phi\_width: 64, phi\_dropout: 0.1, n\_psi\_layers: 2 \\
            &                       & psi\_width: 64, psi\_latent\_width: 128, dot\_prod\_dim: 128, n\_heads: 4 \\
            &                       & attn\_dropout: 0.1, latent\_width: 256, n\_rho\_layers: 2, rho\_width: 512 \\
            &                       & rho\_dropout: 0.1, max\_timescale: 1000., n\_positional\_dims: 8 \\
            \crule
            & Ours                  & weight\_decay: 0.0, n\_train\_latents: 10, n\_train\_samples: 1, n\_test\_latents: 20, n\_test\_samples: 30, \\  %TODO
            &                       & n\_hidden: 128, z\_dim: 32, n\_units: 128, observe\_dropout: 0.3 \\  %TODO
            \midrule
            \multirow{11}{*}{Human Activity}
            & GRU-Mean              & weight\_decay: 0.0, n\_units: 256, dropout: 0.0, recurrent\_dropout: 0.0 \\  %TODO
            & GRU-Simple            & weight\_decay: 0.0, n\_units: 256, dropout: 0.0, recurrent\_dropout: 0.0 \\  %TODO
            & GRU-Forward           & weight\_decay: 0.0, n\_units: 256, dropout: 0.0, recurrent\_dropout: 0.0 \\  %TODO
            & GRU-D                 & weight\_decay: 0.0, n\_units: 256, dropout: 0.0, recurrent\_dropout: 0.0 \\  %TODO
            \crule
            & PhasedLSTM            & weight\_decay: 0.0, n\_units: 256, use\_peepholes: False, leak: 0.01, period\_init\_max: 1000.0 \\
            & Seft                  & weight\_decay: 0.0, n\_phi\_layers: 4, phi\_width: 128, phi\_dropout: 0.2, n\_psi\_layers: 2 \\
            &                       & psi\_width: 64, psi\_latent\_width: 128, dot\_prod\_dim: 128, n\_heads: 4 \\
            &                       & attn\_dropout: 0.5, latent\_width: 32, n\_rho\_layers: 2, rho\_width: 512 \\
            &                       & rho\_dropout: 0.0, max\_timescale: 100., n\_positional\_dims: 4 \\
            \crule
            & Ours                  & weight\_decay: 0.0, n\_train\_latents: 10, n\_train\_samples: 1, n\_test\_latents: 10, n\_test\_samples: 10, \\  %TODO
            &                       & n\_hidden: 128, z\_dim: 10, n\_units: 128, observe\_dropout: 0.2 \\  %TODO
            \bottomrule
        \end{tabular}
    }
    \label{tab:hyperparam_classification}
\end{table*}

\subsubsection{Evaluation Metrics}
For the classification task, we evaluate all models in terms of both predictive accuracy and predictive uncertainty.
We use the area under receiver operating characteristic (AUROC), and the accuracy (ACC) to evaluate the predictive performance. To measure the uncertainty calibration of the model, we use cross entropy (CE), expected calibration error (ECE), and brier score (BS) for comparing calibration.

% In addition, we also check the balanced versions of uncertainty metrics due to severe class imbalance of datasets.

% \paragraph{Balanced metric}
% In supervised dataset $\calD$, which contains input data $\bsx$ and a corresponding label $y$, we simply re-weight each uncertainty metric by class ratio.
% \[
%     M_\text{bal}(\calD) = \frac{1}{C} \sum_{c} M(\calD_c)
% \]
% Here, $\calD_c$ is a subset of the dataset $\calD$ which only contains the label $y = c$.

\paragraph{Accuracy Metrics}

Accuracy metrics are defined using the following terms, where $tp$, $tn$, $fn$, and $fp$ denote true positive, true negative, false negative, and false positive respectively.

\[
    \text{accuracy} = \frac{tp + tn}{tp + fp + fn + tn} \\
    \text{precision} = \frac{tp}{tp + fp} \\
    \text{recall} = \frac{tp}{tp + fn} \\
    \text{sensitivity} = \frac{tp}{tp + fn}
\]

\begin{itemize}
    % \item \textbf{AUPRC}~\citep{schutze2008introduction}: area under precision recall curve.
    \item \textbf{AUROC}: area under receiver operating characteristic, area under sensitivity curve.
\end{itemize}

\subsection{Details for imputation experiments}
Basically, We randomly delete 10\% of observed data for testing the imputation performance of models. We trained our model with five different seeds and also measured the performance of five different seeds.

\subsubsection{Baseline methods}
\begin{itemize}
    \item \textbf{Mean}: Replace missing values with global mean.
    \item \textbf{Forward}: Impute missing value with previously observed value
\item \textbf{GP-VAE}: This model is VAE-based probabilistic imputation method proposed by \citet{forutin2020gpvae}. This method employs GP-prior to encode the temporal correlation in the latent space.
     \item \textbf{SAITS}: This model is self-attention based imputation model which \citet{du2022saits} proposed. This employs a combination of multiple layers of transformer encoder to impute multivariate time series data 
     \item \textbf{Ours w/o supervision}: To analyze the effect of supervised signal to the imputation performance, we remove the supervised term from our training objective and train only the generative part of our architecture.
     \item \textbf{Ours w/o dropout}: Our model without obsdropout.
    
    \item \textbf{Ours w/o supervision and MNAR}: Our model without obsdropout and MNAR assumption, which missing model and classifier is removed from our model.

\end{itemize}

\subsubsection{Evaluation Metrics}
For the imputation task, we evaluate all methods in terms of MRE(Mean Relative Error) and MAE(Mean Absolute Error). 

\subsubsection{Training details}
We applied early stopping with patience 20 and stopping criterion as validation loss of each model. We randomly erase 10\% of the observed data for all datasets and measured the evaluation metrics. 
We use almost the same hyperparameter settings as in the classification tasks.

% \subsubsection{Hyperparameters}

% See Table~\ref{tab:hyperparam_imputation} for detailed hyperparameter settings for the imputation experiment.
% \input{table/hyperparam_imputation}

% \input{appendix/additional}
%%%%%%%%%%%%%%%%%%%%%%%%%%%%%%%%%%%%%%%%%%%%%%%%%%%%%%%%%%%%%%%%%%%%%%%%%%%%%%%
%%%%%%%%%%%%%%%%%%%%%%%%%%%%%%%%%%%%%%%%%%%%%%%%%%%%%%%%%%%%%%%%%%%%%%%%%%%%%%%

\end{document}